\documentclass[10pt,twocolumn,letterpaper]{article}

\usepackage{cvpr}
\usepackage{times}
\usepackage{epsfig}
\usepackage{graphicx}
\usepackage{amsmath}
\usepackage{amssymb}
\usepackage{algorithm}
\usepackage{algorithmic}

\usepackage[pagebackref=true,breaklinks=true,letterpaper=true,colorlinks,bookmarks=false]{hyperref}
 \cvprfinalcopy 
\def\cvprPaperID{2388} 

\begin{document}

\title{Range Loss for Deep Face Recognition with Long-tail}

\author{
Xiao Zhang$^{1,4}$, Zhiyuan Fang$^{2,4}$, Yandong Wen$^3$, Zhifeng Li$^{4}$, Yu Qiao$^{4}$\\
\small $^1$The School of Computer Software, Tianjin University\\
\small $^2$Dept. of Computer Science, Southern University of Science and Technology\\
\small $^3$Dept. of Electrical and Computer Engineering, Carnegie Mellon University\\
\small $^4$Shenzhen Key Lab of Comp. Vis. \& Pat. Rec., Shenzhen Institutes of Advanced Technology, CAS\\
    \tt\small zhangxiao1688@tju.edu.cn fangzy@mail.sustc.edu.cn yandongw@andrew.cmu.edu\\
    \tt\small \{zhifeng.li, yu.qiao\}@siat.ac.cn
}

\maketitle

\begin{abstract}
Convolutional neural networks have achieved great improvement on face recognition in recent years because of its extraordinary ability in learning discriminative features of people with different identities. To train such a well-designed deep network, tremendous amounts of data is indispensable. Long tail distribution specifically refers to the fact that a small number of generic entities appear frequently while other objects far less existing. Considering the existence of long tail distribution of the real world data, large but uniform distributed data are usually hard to retrieve. Empirical experiences and analysis show that classes with more samples will pose greater impact on the feature learning process\cite{zhou2015naive, ouyang2016factors} and inversely cripple the whole models feature extracting ability on tail part data.
Contrary to most of the existing works that alleviate this problem by simply cutting the tailed data for uniform distributions across the classes, this paper proposes a new loss function called range loss to effectively utilize the whole long tailed data in training process.
More specifically, range loss is designed to reduce overall intra-personal variations while enlarging inter-personal differences within one mini-batch simultaneously when facing even extremely unbalanced data.
The optimization objective of range loss is the $k$ greatest range's harmonic mean values in one class and the shortest inter-class distance within one batch.
Extensive experiments on two famous and challenging face recognition benchmarks (Labeled Faces in the Wild (LFW)\cite{LFWTech} and YouTube Faces (YTF)\cite{wolf2011face}) not only demonstrate the effectiveness of the proposed approach in overcoming the long tail effect but also show the good generalization ability of the proposed approach.
\end{abstract}

\section{Introduction}
Convolutional neural networks (CNNs) have witnessed great improvement on a series of vision tasks such as object classification \cite{krizhevsky2012imagenet,simonyan2014very, szegedy2015going, he2015delving, he2015deep} , scene understanding \cite{zhou2014learning, zhou2014object}, and action recognition \cite{karpathy2014large}. As for the face recognition task, CNNs like DeepID2+ \cite{sun2014deep} by Yi Sun, FaceNet\cite{schroff2015facenet}, DeepFace\cite{taigman2014deepface}, Deep FR\cite{parkhi2015deep}, have even proven to outperform humans on some benchmarks.

To train a robust deep model, abundant training data \cite{deng2009imagenet} and well-designed training strategies are indispensable. It is also worth to point out that, most of the existing training data sets like LSVRC's object detection task \cite{ILSVRC15}, which contains 200 basic-level categories, were carefully filtered so that  the number of each object instance is kept similar to avoid the long tailed distribution.

\begin{figure}
\includegraphics[width=1.0\linewidth]
               {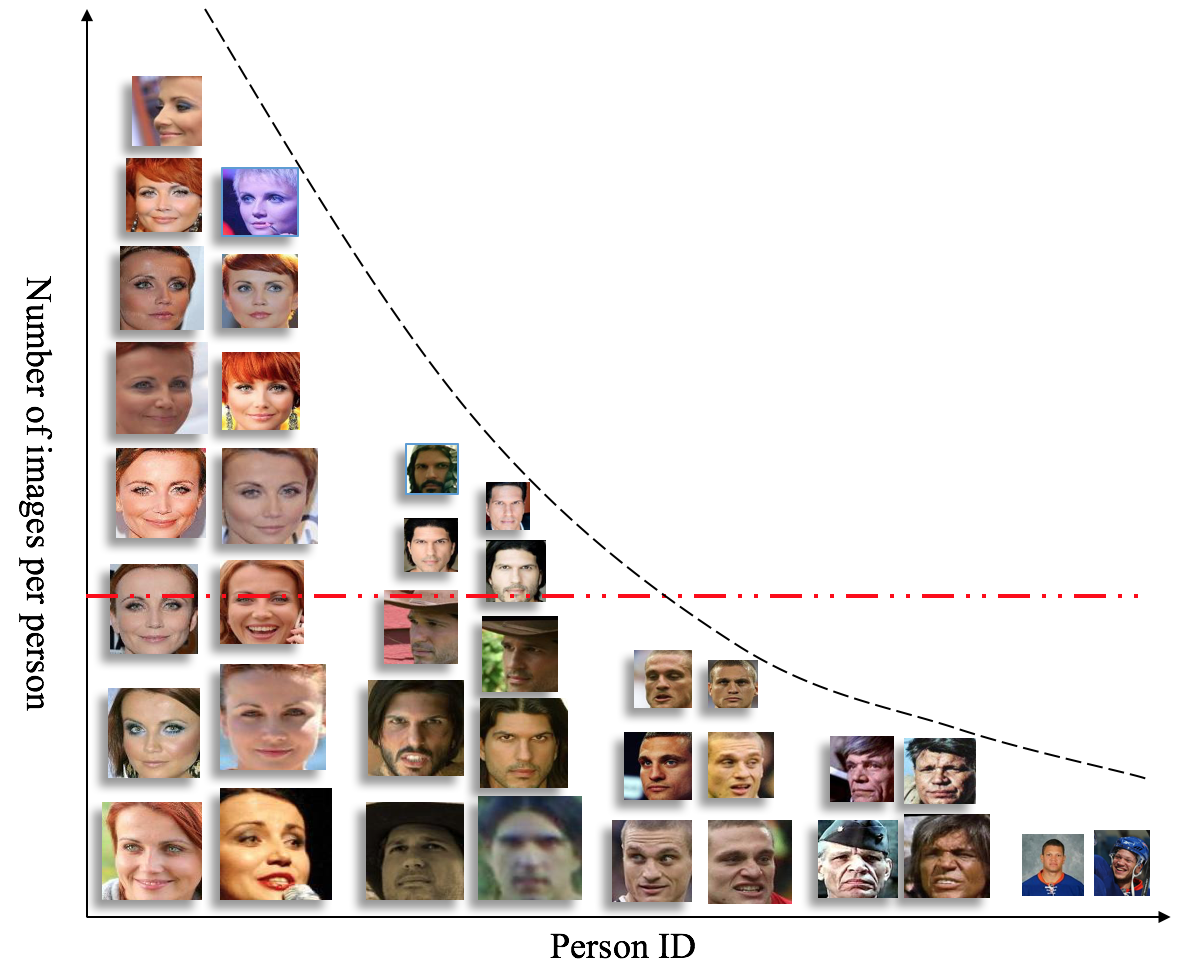}
\caption{Long tail distributed data set for human faces(Selected from MS-Celeb-1M\cite{guo2016ms}). Number of face images per person falls drastically, and only a small part of persons have large number of images. Cutting line in red represents the average number of images per person.}
\label{fig:overview}
\end{figure}

More specifically, long tail property refers to the condition where only limited number of object classes appear frequently, while most of the others remain relatively rarely. If a model was trained under such an extremely imbalanced distributed dataset (in which only limited and deficient training samples are available for most of the classes), it would be very difficult to obtain good performance. In other words, insufficient samples in poor classes/identities will result in the intra-class dispension in a relatively large and loose area, and in the same time compact the inter-classes dispension\cite{wen2016discriminative}.

In \cite{againstlongtail}, Bengio gave the terminology called ``representation sharing'': human possess the ability to  recognize objects we have seen only once or even never as  representation sharing. Poor classes can be beneficial for knowledge learned from semantically similar but richer classes.
While in practice, other than learning the transfer feature from richer classes, previous work mainly cut or simply replicate some of the data to avoid the potential risk long tailed distribution may cause. According to \cite{ouyang2016factors}'s verification, even only 40\% of positive samples are left out for feature learning, detection performance will be improved a bit if the samples are more uniform. Such disposal method's flaw is obvious: To simply abandon the data partially, information contained in these identities may also be omitted.

In this paper, we propose a new loss function, namely range loss to effectively enhance the model's learning ability towards tailed data/classes/identities. Specifically, this loss identifies the maximum Euclidean distance between all sample pairs as the range of this class. During the iteration of training process, we aim to minimize the range of each class within one batch and recompute the new range of this subspace simultaneously.


The main contributions of this paper can be summarized as follows:

1. We extensively investigate the long tail effect in deep face recognition, and propose a new loss function called range loss to overcome this problem in deep face recognition. To the best of our knowledge, this is the first work in the literature to discuss and address this important problem.

2. Extensive experiments have demonstrated the effectiveness of our new loss function in overcoming the long tail effect. We further demonstrate the excellent generalizability of our new method on two famous face recognition benchmarks (LFW and YTF).

\section{Related Work}

Deep learning is proved to own a great ability of feature learning and achieve great performances in a series of vision tasks like object detection \cite{gupta2014learning, sermanet2013overfeat, lin2013network, he2014spatial, szegedy2015going}, face recognition \cite{parkhi2015deep, schroff2015facenet, sun2014deep, chen2013blessing, yang2014context, norouzi2013zero, wen2016latent}, and so forth. By increasing the depth of the deep model to 16-19 layers, VGG \cite{simonyan2014very} achieved a significant improvement on the VOC 2012 \cite{Everingham10} and Caltech 256 \cite{griffin2007caltech}. Based on the previous work, Residual Network, proposed by Kaiming He \textit{et al}, present a residual learning framework to ease the training of substantially deeper networks \cite{he2015deep}.
In \cite{wen2016discriminative}, the authors propose a new supervision signal, called center loss, for face recognition task. Similar to our range loss's main practice, center loss minimizes the distances between the deep features and their corresponding class centers ( Defined as arithmetic mean values).

Long tailed distribution of the data has been involved and studied in scene parsing \cite{yang2014context}, and zero-shot learning \cite{norouzi2013zero}.
In a workshop talk 2015, Bengio described the long tail distribution as the enemy of machine learning\cite{againstlongtail}. In \cite{yang2014context}, a much better super-pixel classification results are achieved by the expanding the poor classes' samples.
In \cite{ouyang2016factors}, this paper investigates many factors that influence the performance in fine-tune for object detection with long tailed distribution of samples. Their analysis and empirical results indicate that classes with more samples will pose greater impact on the feature learning. And it is better to make the sample number more uniform across classes.
\section{The Proposed Approach}
In this section, we firstly elaborate our exploratory experiments implemented with VGG on LFW's face verification task, which give us an intuitive understanding of the potential effects by long tailed data. Based on the conclusion drew from these two experiments, we propose a new loss function namely, range loss to improve model's endurance and utilization rate toward highly imbalanced data follow by some discussions.

\subsection{Problem formulation}

\begin{figure}[t]
   \includegraphics[width=1.0\linewidth]{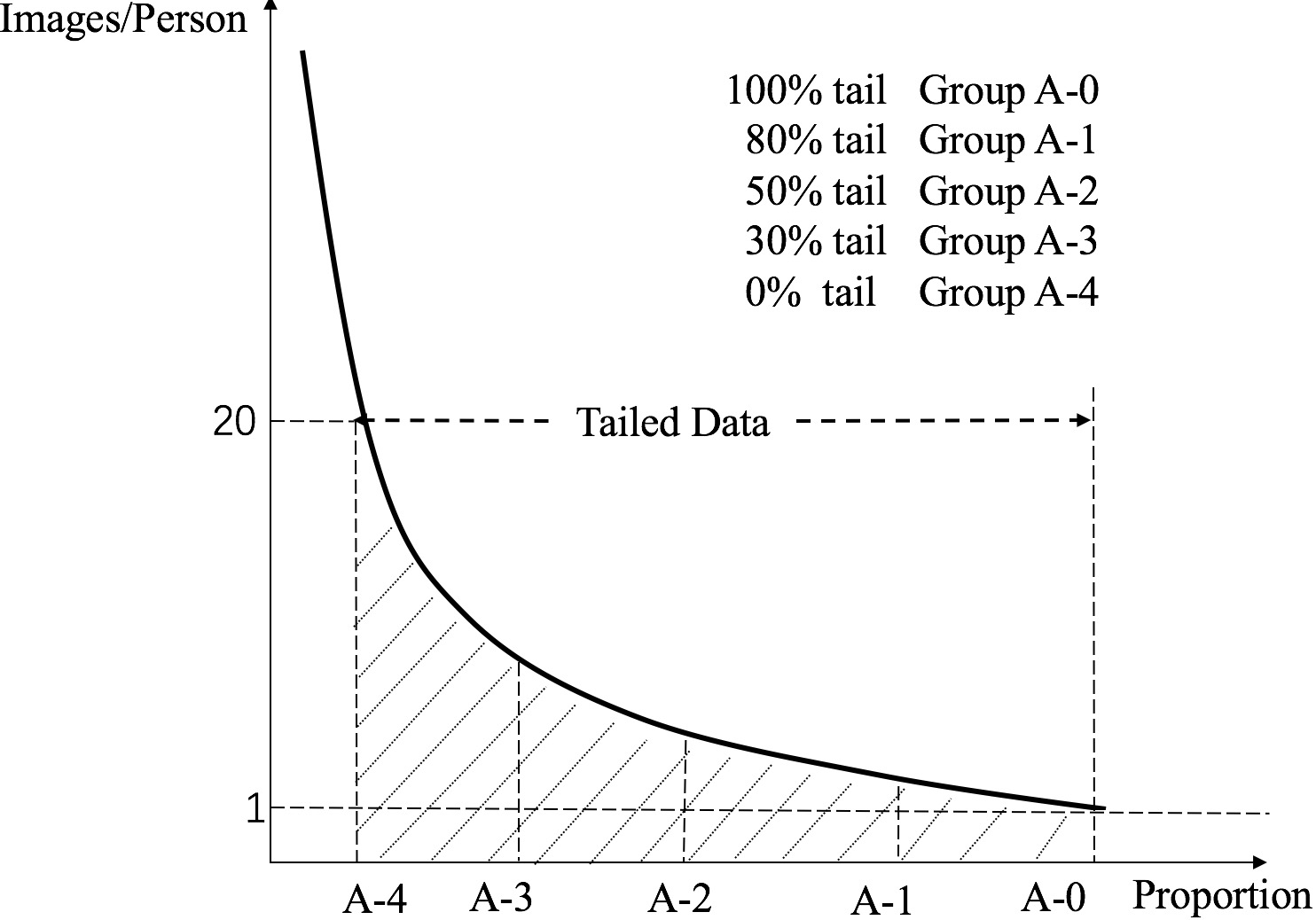}
   \caption{ Our Constructed Data set with Long-tailed Distributions. The Cutting lines in the above figure represent the division proportions we used to construct  subsets of object classes.}
\label{fig:long}
\end{figure}

In statistics, a long tail of certain distributions is the portion of the distribution having a large number of occurrences far from the "head" or central part of the distribution \cite{Bingham2011The}.
To investigate the long-tail property deeply and thoroughly in the context of deep learning face recognition, we first trained several VGG-16 models \cite{simonyan2014very} with softmax loss function on data sets with extremely imbalanced distribution ( the distribution of our training data is illustrated in \ref{fig:long}. )
We constructed our long tail distributed training set from MS-Celeb-1M \cite{guo2016ms} and CASIA- WebFace\cite{yi2014learning} data set, which consists of 1.7 million face images with almost 100k identities included in the training data set. Among this set, there are 700k images for roughly 10k of the identities, and 1 million images for the remaining 90k identities. To better understand the potential effect of long tailed data on the extracted identical representation features, we slice the raw data into several groups according to different proportions in Table~\ref{table:traindata}. As we can see in Fig \ref{fig:long}, classes that contain less than 20 images are defined as poor classes (tailed data).
As is shown in Table1, group A-0 is the raw training set.  20\%, 50\%, 70\%, 100\% of the poor classes in A-0 is cut to construct group A-1, A-2, A-3 and A-4 respectively.
\begin{table}
\begin{center}

\begin{tabular}{|c|c|c|c|}
\hline
 Groups &Num of Identities & Images & Division Ratio \\
\hline\hline
A-0 & 99,891 & 1,687,691 & 0.00\% \\
A-1 & 81,913 & 1,620,526 & 20.00\% \\
A-2 & 54,946 & 1,396,414 & 50.00\% \\
A-3 & 26,967 & 1,010,735 & 70.00\% \\
A-4 & 10,000 & 699,832	 & 100.00\% \\
\hline
\end{tabular}
\end{center}
\caption{Training Set with Long-tail Distribution. Control group's division proportion can be viewed in Fig. \ref{fig:long}}
\label{table:traindata}
\end{table}
We conduct our experiments on LFW's face verification task and the accuracy are compared in Table~\ref{table:resultlongtail}.
As is shown in Table~\ref{table:resultlongtail}, group A-2 achieves the highest accuracy rate in series A. With the growth of the tail, group A-1 and A-0 get lower performances though they contain more identities and images.

These results indicate that, tailed data stand a great chance to pose a negative effect on the trained model's ability.
Based on the above findings, we come to analyze the distinct characteristics of Long-tail effect that, conventional visual deep models do not always benefit as much from larger data set with long-tailed property as it does for a uniform distributed larger data set. Moreover, long tailed data set, if cut and remained in a specific proportion (50\% in here), will contribute to deep models' training.

In fact, there are some different features in face recognition task: the intra-class variation is large because the face image can be easily influenced by the facing directions, lighting conditions and original resolutions. On the other hand, compared with other recognition tasks, the inter class variation in face recognition is much smaller. As the growth of the number of identities, it is possible to include two identities with similar face. Worse still, their face images are so few that can not give a good description to their own identities.


\begin{table}
\begin{center}
\begin{tabular}{|c|c|}
\hline
Groups & Acc. on LFW \\
\hline\hline
A-0 (with long-tail)& 97.87\% \\
A-1 (cut 20\% tail)& 98.03\% \\
A-2 (cut 50\% tail)& 98.25\% \\
A-3 (cut 70\% tail)& 97.18\% \\
A-4 (cut 100\% tail)& 95.97\% \\

\hline
\end{tabular}
\end{center}
\caption{VGG Net with Softmax Loss's performances on LFW with Long-tail Effect.}
\label{table:resultlongtail}
\end{table}

\subsection{Study of VGG Net with Contrastive and Triplet Loss on Subsets of Object Classes}
Considering the characteristics of long tailed distributions: a small number of generic objects/entities appear very often while most others exist much more rarely. People will naturally think the possibility to utilize the contrastive loss\cite{sun2014deep} or the triplet loss\cite{schroff2015facenet} to solve the long tail effect because of its pair training strategy.

The contrastive loss function consists of two types of samples: positive samples of similar pairs and negative samples of dissimilar pairs. The gradients of the loss function act like a force that pulls together positive pairs and pushes apart in negative pairs. Triplet loss minimizes the distance between an anchor and a positive sample, both of which have the same identity, and maximizes the distance between the anchor and a negative of a different identity.

In this section, we apply the contrastive loss and triplet loss on VGG-16 with the same constructed long tailed distributed data. The goal of this experiment, on some level, is to gain insights on the contrastive loss and triplet loss's processing capacity of long tailed data. We conduct the LFW's face verification experiment on the most representative groups A-0 and group A-2 with full and half of the long tailed data.
As for the training pairs, we depart all identities into two parts with same number of identities firstly.
The former part contains only richer classes and the later poor classes.
Positive pairs (images of the same person) are randomly selected from the former part and negative pairs are generated in the latter part data of different identities. After training, we got the contrastive and triplet's results shown in Table \ref{contrastiveresult} and Table \ref{table:tripletresult} respectively.
From these tables, we can clearly see that long tail effect still exist on models trained with contrastive loss and triplet loss: with 291,277 more tailed images in group A-0's training set, contrary to promoting the verification performances, accuracy is reduced by 0.15\%. Moreover, contrastive loss improves the accuracy by 0.46\% and  0.21\% comparing to VGG-16 with softmax loss.

Probable causes of long tail effect's existence in contrastive loss may lie that:
though pair training and triplet training strategy can avoid the direct negative effect long tail distribution may brought, classes in the tail are more like to be selected in the training pairs' construction (poor classes are accounted for 90\% of the classes). Because the massive classes with rare samples piled up in the tail, pairs contain the  pictures of one person are extremely limited in a small amount, thus resulting in the lack of enough descriptions toward intra-class's invariation.
Inspired by contrastive and triplet loss's defect and deficiency, we find the necessity to propose our loss function specially-costumed to be integrated into training data with long tail distribution. Such loss function is designed primarily for better utilizing the tailed data, which we believe has been submerged by the richer classes' information and poses not only almost zero impact to the model, but a negative resistance to model's effectiveness in learning  discriminative features.

\begin{table}
\begin{center}
\begin{tabular}{|c|c|}
\hline
Training Groups & Acc. on LFW  \\
\hline\hline
A-0 (with long-tail)& 98.35\%\\
A-2 (cut 50\% of tail)& 98.47\%\\
\hline
\end{tabular}
\end{center}
\caption{VGG Net with Softmax+Contrastive Loss's performances on LFW with Long-tail Effect.}
\label{contrastiveresult}
\end{table}

\begin{table}
\begin{center}
\begin{tabular}{|c|c|}
\hline
Training Groups & Acc. on LFW \\
\hline\hline
A-0 (with long-tail)& 98.10\% \\
A-2 (cut 50\% of tail)& 98.40\% \\
\hline
\end{tabular}
\end{center}
\caption{VGG Net with Softmax+Triplet Loss's performances on LFW with Long-tail Effect.}
\label{table:tripletresult}
\end{table}

\subsection{The Range Loss}

\begin{figure}
\includegraphics[width=1.0\linewidth]
               {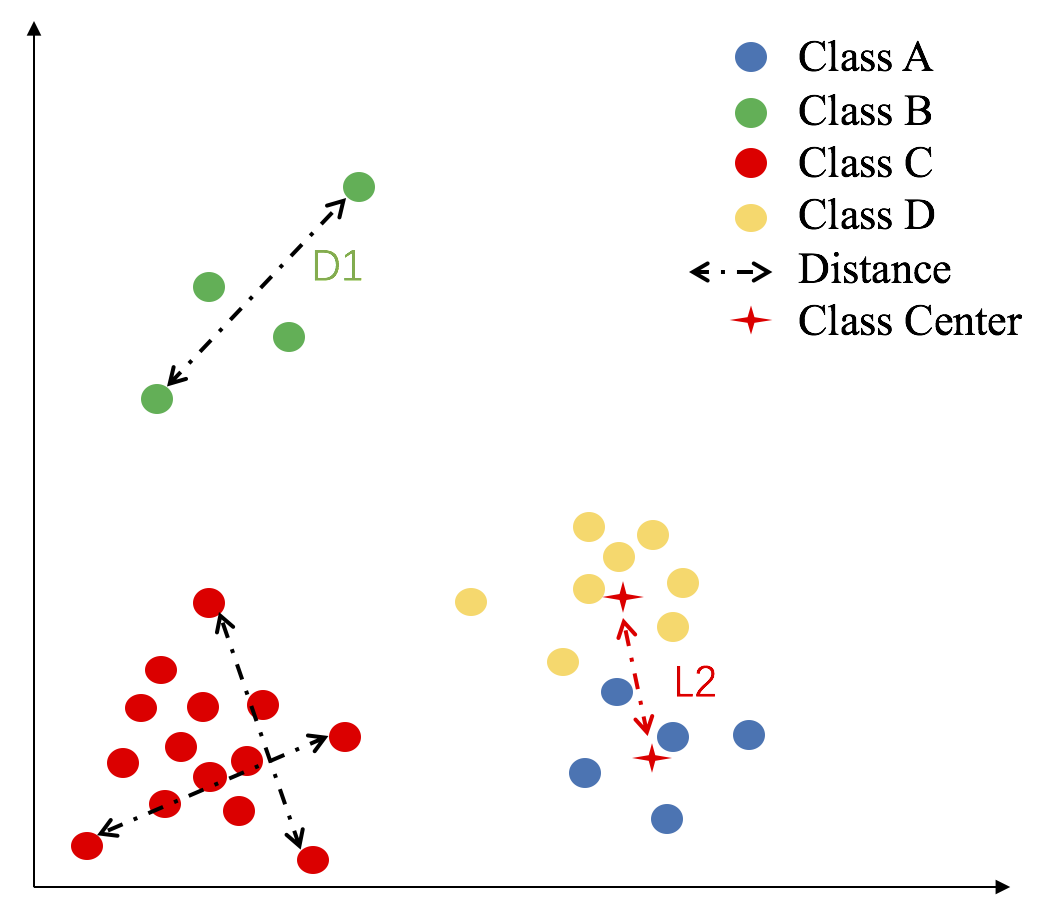}
\caption{An simulated 2-D feature distribution graph in one mini-batch. There are 4 classes in this mini-batch, and Class B represents one typical poor class. $D_1$ denotes Class B's greatest intra-class range. $L_2$ between Class D and Class A represents the center distance of these two classes. The objective of range loss can be seen as the shortest center distances( $L_2$ in these 4 classes) and the harmonic mean value of the $k$ greatest ranges( $D_1$ as for Class B) in each class. (Best viewed in color.) }
\label{fig:distance}
\end{figure}

Intrigued by the experiment results above that long tail effect does exist in models trained with contrastive loss and triplet loss, we delve deeper into this phenomenon, give a qualitative explanation of the necessity to propose our new loss toward this problem and further discuss the merits and disadvantages of the existing methods.

In long tail distributed data, samples of the tailed data are usually extremely rare, there are only very limited images for each person in our dataset. Contrastive loss optimizes the model in such a way that neighbors are pulled together and non-neighbors are pushed apart. To construct such a training set consists of similar pairs and negative examples of dissimilar pairs, sufficient pairs of the same person is indispensable but out of the question to be achieved on long tailed data.

Moreover, as we discussed in the previous section, richer classes will pose greater impact on the model's training. Ways to leverage the imbalanced data should be considered.

The the objective of designing range loss is summarized as:
\begin{itemize}
  \item Range loss should be able to strengthen the tailed data's impact in the training process to prevent poor classes from being submerged by the rich classes.
  \item Range loss should penalize those sparse samples' dispension brought by poor classes.
  \item Enlarge the inter-class distance at the same time.
  \end{itemize}

Inspired by the contrastive loss, we design the Range Loss in a form that reduces intra-personal variations while enlarge the inter-personal differences simultaneously.
But contrary to contrastive loss function's optimizing on positive and negative pairs,
the range loss function will calculate gradients and do back propagation based on the overall distance of classes within one mini－batch. In other words, statistical value over the whole class substituted the single sample's value on pairs.
As to the second goal, the author in \cite{henriques2013beyond} use the hard negative mining idea to deal with these samples.
For those sparse training samples in poor classes, features located in the feature space's spatial edge(edge feature) can be viewed as the points that enlarge the intra-class's invariation most. These samples, to a certain degree, can also be viewed as the hard negative samples. Inspired by this idea, range loss should be designed to minimize those hard negative samples' distance thus lessen the exaggerated intra-class invariation by tailed data. Based on this, we calculate $k$ greatest range's harmonic mean value over the feature set extracted in the last FC layer
as the inter-class loss in our function. The range value can be viewed as the intra-class's two most hard negative samples. For the inter-class loss, the shortest distance of class feature centers will be the supervision.

To be more specifically, range loss can be formulated as:

\begin{eqnarray}
\mathcal{L}_{R}=\alpha \mathcal{L}_{R_{intra}}+\beta \mathcal{L}_{R_{inter}}
\end{eqnarray}
Where $\alpha$ and $\beta$ are two weight of range loss and in which $\mathcal{L}_{R_{intra}}$ denotes the intra-class loss that penalizes the maximum harmonic range of each class:
\begin{eqnarray}
\begin{split}
\mathcal{L}_{R_{intra}}=\sum_{i\subseteq I} \mathcal{L}_{R_{intra}}^{i}
=\sum_{i\subseteq I} \frac{k}{\sum_{j=1}^{k}\frac{1}{\mathcal{D}_{j}}}
\end{split}
\end{eqnarray}

Where $I$ denotes the complete set of classes/identities in this mini-batch. $\mathcal{D}_{j}$ is the $j$-th largest distance. For example, we define $\mathcal{D}_{{1}} = \left\|x_{{1}}-x_{{2}}\right\|_{2}^{2}$
and $\mathcal{D}_{{2}} = \left\|x_{{3}}-x_{{4}}\right\|_{2}^{2}$.
 $\mathcal{D}_{{1}}$ and $\mathcal{D}_{{2}}$ are the largest and
second largest Euclidean range for a specific identity $i$ respectively. Input $x_{{1}}$
and $x_{{2}}$ denoted two face samples with the longest distance, and similarly, input $x_{{3}}$
and $x_{{4}}$ are samples with of the second longest distance. Equivalently, the overall cost is the
harmonic mean of the first $k$-largest range within each class. Experience shows that $k=2$ bring a good performance.

 $\mathcal{L}_{R_{inter}}$ represents the inter-class loss that
\begin{eqnarray}
\begin{split}
\mathcal{L}_{R_{inter}}=\max(m-\mathcal{D}_{Center}, 0)\\=\max(m-\left \| \overline{x}_\mathcal{Q}-\overline{x}_\mathcal{R} \right \|_{2}^{2}, 0)
\end{split}
\end{eqnarray}

where, $\mathcal{D}_{Center}$ is the shortest distance between class centers, that are defined as the arithmetic mean of all output features in this class. In a mini-batch, the distance between the center of class $\mathcal{Q}$ and class $\mathcal{R}$ is the shortest distance for all class centers. $m$ denotes a super parameter as the max optimization margin that will exclude $\mathcal{D}_{Center}$
greater than this margin from the computation of the loss.

In order to prevent the loss being degraded to zeros \cite{wen2016discriminative} during the training, we use our loss joint with the softmax loss as the supervisory signals.
The final loss function can be formulated as:

\begin{eqnarray}
\begin{split}
\mathcal{L}=\mathcal{L}_{M}+\lambda \mathcal{L}_{R}
=-\sum_{i=1}^{M} \log \frac{e^{W_{y_{i}}^{T}x_{i}+b_{y_{i}}}}{\sum_{j=1}^{n} e^{W_{j}^{T}x_{i}+b_{j}}}+\lambda \mathcal{L}_{R}
\end{split}
\end{eqnarray}

In the above expression, $M$ refers to the mini-batch size and $n$ is the number of identities within the training set. $x_{i}$ denotes the features of identity $y_{i}$ extracted from our deep model's last fully connected layers. $W_{j}$ and $b_{j}$ are the parameters of the last FC layer. $\lambda$ is inserted as a scaler to balance the two supervisions. If set to 0, the overall loss function can be seen as the conventional softmax loss.
According to the chain rule, gradients of the range loss with respect to $x_i$ can be computed as:
\begin{eqnarray}
\begin{split}
\frac{\partial \mathcal{L}_{R}}{\partial x_i}
=\alpha \frac{\partial \mathcal{L}_{R_{intra}}}{\partial x_i}+\beta \frac{\partial \mathcal{L}_{R_{inter}}}{\partial x_i}
\end{split}
\end{eqnarray}


For a specific identity, let $S=\sum_{i=1}^{k}\frac{1}{\mathcal{D}_{i}}$, $D_j$ is a distance of $x_{j1}$ and $x_{j2}$, two features in the identity.
\begin{eqnarray}
\begin{split}
\frac{\partial \mathcal{L}_{R_{intra}}}{\partial x_{i}}=\frac{2k}{(D_{j}S)^2}
\left\{
\begin{matrix}
\left | x_{j1}-x_{j2} \right |, x_i=x_{j1}\\
\left | x_{j2}-x_{j1} \right |, x_i=x_{j2}\\
0, x_i \neq x_{j1}, x_{j2}
\end{matrix}\right.
\end{split}
\end{eqnarray}

\begin{eqnarray}
\begin{split}
 \frac{\partial \mathcal{L}_{R_{inter}}}{\partial x_{i}}
=\left\{
\begin{matrix}
\frac{\partial \mathcal{L}}{\partial x_ \mathcal{Q}}=\frac{1}{2n_R}\left | \frac{\sum x_R}{n_R}-\frac{\sum x_\mathcal{Q}}{n_\mathcal{Q}} \right |\\
\frac{\partial  \mathcal{L}}{\partial x_R}=\frac{1}{2n_\mathcal{Q}}\left | \frac{\sum x_\mathcal{Q}}{n_\mathcal{Q}}-\frac{\sum x_R}{n_R} \right |\\
0, x_i\neq x_Q, x_R
\end{matrix}\right.
\end{split}
\end{eqnarray}

Where $n_i$ denotes the total number of samples in class $i$. And we summarize the loss value and gradient value's computation process in Algorithm 1.
 \ref{fig:distance}).

\begin{algorithm}
  \caption{Training algorithm with range loss}
  \label{alg1}
  \begin{algorithmic}
  \REQUIRE Feature set $\{x_i\}$ extracted from the last fully connected layer. Hyper parameter $m$ and $ \lambda$.
  \ENSURE The intra-class part of range loss $L_{R_{intra}}$ and the inter-class part of range loss $L_{R_{inter}}$. The gradient of intra-class $\frac{\partial L_{R_{intra}}}{\partial x_{i}}$ and inter-class $ \frac{\partial L_{R_{inter}}}{\partial x_{i}}$.
  \STATE
  \FOR{each class ${i \subseteq I}$ in one mini-batch}
  \STATE Compute the arithmetic mean feature as feature center $c_i$ of class $i$.
  \STATE Compute the $k$ largest Euclidean distances $\{D_j\}$ among features $\{x_i\}$ of class $i$.
  \STATE Compute the harmonic mean of $\{D_j\}$ as the intra-class loss of class $i$, $L^i_{R}=\frac{k}{\sum_{j=1}^{k}D_{j}}$.
  \ENDFOR
  \STATE Compute the intra-class loss $L_{R_{intra}}=\sum_{i \subseteq I}L^i_{R}=\sum_i \frac{k}{\sum_{j=1}^{k}D_{j}}$.
  \STATE Compute the intra-class gradient $\frac{\partial L_{R_{intra}}}{\partial x_{i}}$.
  \STATE Compute the shortest distances $D_{center}$ among all  feature centers $\{ c_P\}$.
  \IF{$m-D_{min} > 0$}
  \STATE Output the inter-class gradient $ \frac{\partial L_{R_{inter}}}{\partial x_{i}}$.
  \ELSE
  \STATE  $ \frac{\partial L_{R_{inter}}}{\partial x_{i}}=0$.

  \ENDIF
  \end{algorithmic}
\end{algorithm}

\begin{figure}[t]

\includegraphics[width=1.0\linewidth]
               {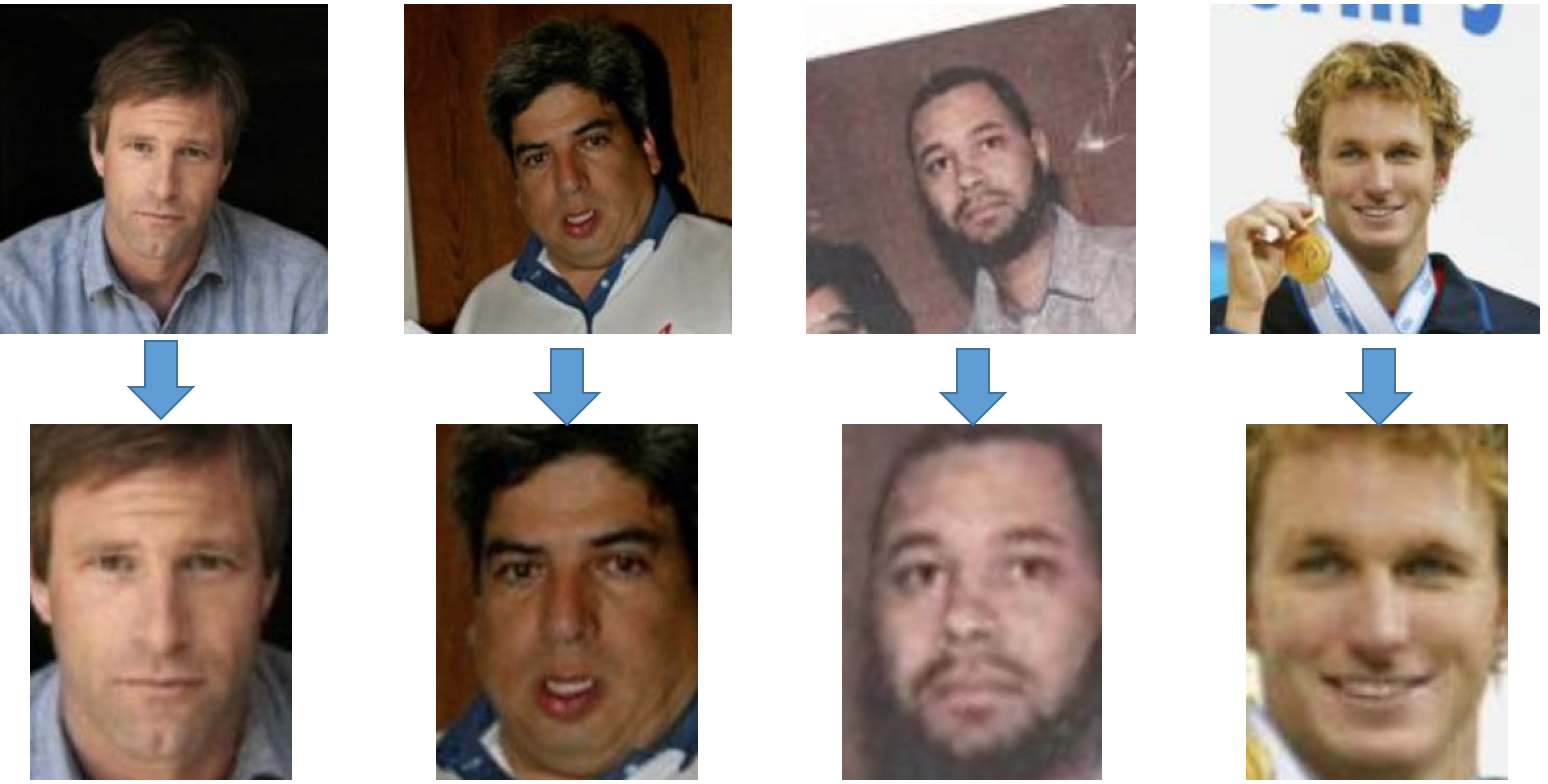}
\caption{Some common face images in LFW.}
\label{fig:lfw}
\end{figure}

\begin{figure}[t]

\includegraphics[width=1.0\linewidth]
               {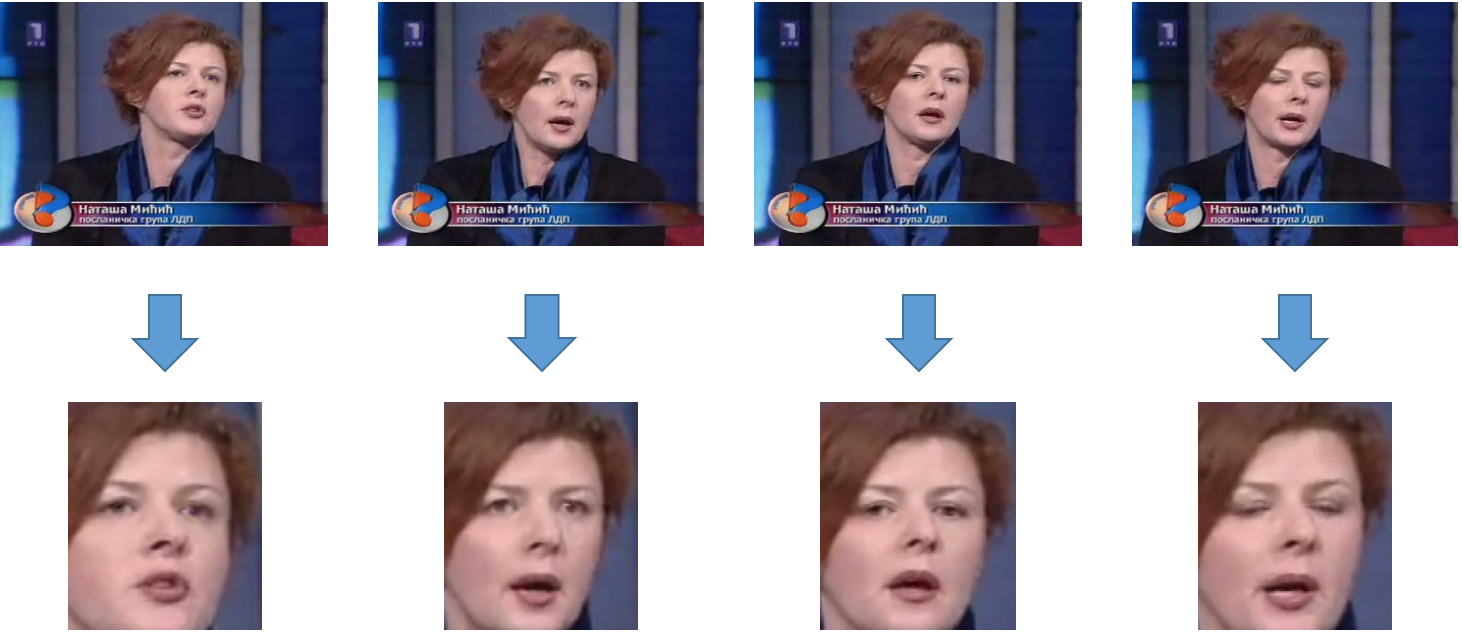}
\caption{Some common face images in YTF.}
\label{fig:lfw}
\end{figure}

\begin{figure*}[h]

\includegraphics[width=1.0\linewidth]
               {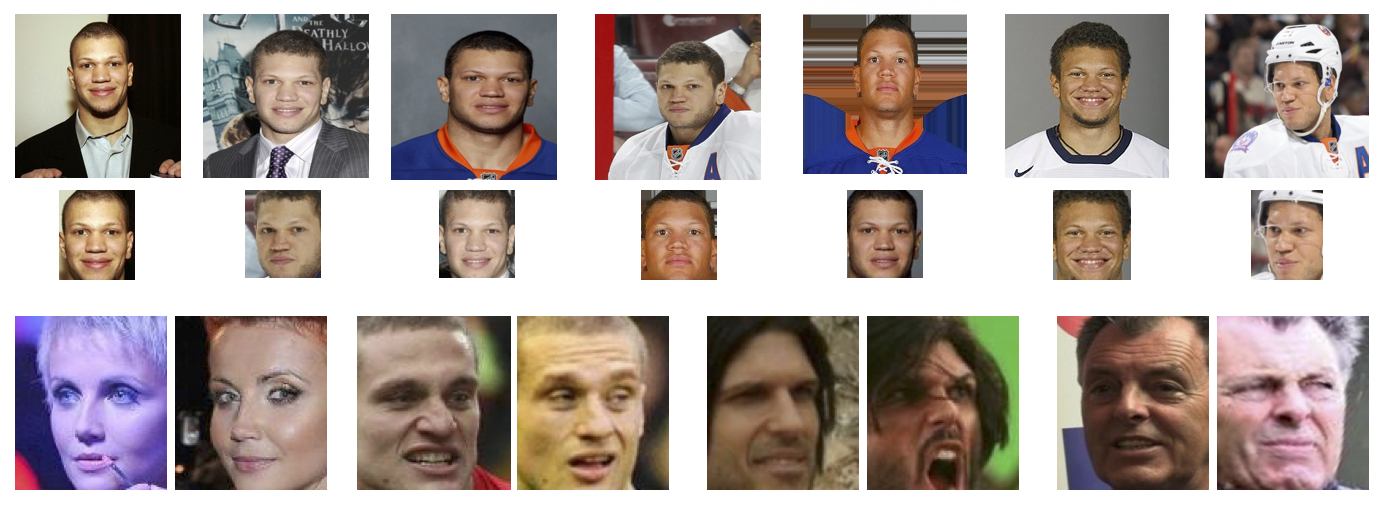}
\caption{An Overview of Our Filtered and Cropped Face Database. Images in the first row are raw images before alignment and cropping. Corresponding images are listed below the raw images. Some common faces in our training set are presented in the last row.  }
\label{fig:over}
\end{figure*}

\subsection{Discussions on Range Loss's Effectiveness}
Generally speaking, range loss adopts two stronger identifiability statistical parameters than contrastive loss and others: distance of the peripheral points in the intra-class subspace, and the center distance of the classes. Both the range value and the center value is calculated based on groups of samples.  Statistically speaking, range loss utilizes those training samples of one mini-batch in a joint way instead of individually or pairly, thus ensure the model's optimization direction comparatively balanced. To give an intuitive explanations of the range loss, we have simulated a 2-D feature distribution graph in one mini-batch with 4 classes (see Fig. \ref{fig:distance})

\section{Experiments}
In this section, we evaluate our range loss based models on two well known face recognition benchmarks, LFW and YTF data sets. We firstly implemented our range loss with VGG's \cite{simonyan2014very} architecture and train on 50\% and 100\%  long tailed data to measure its performances on face verification task. More than that, based on \cite{wen2016discriminative}'s recent proposed center loss which achieves the state-of-art performances on LFW and YTF, we implement our range loss with the same network's structure to see whether the range loss is able to handle the long tailed data better than other loss function in a more general CNN's structure.

\begin{figure*}

\includegraphics[width=1.0\linewidth]
                {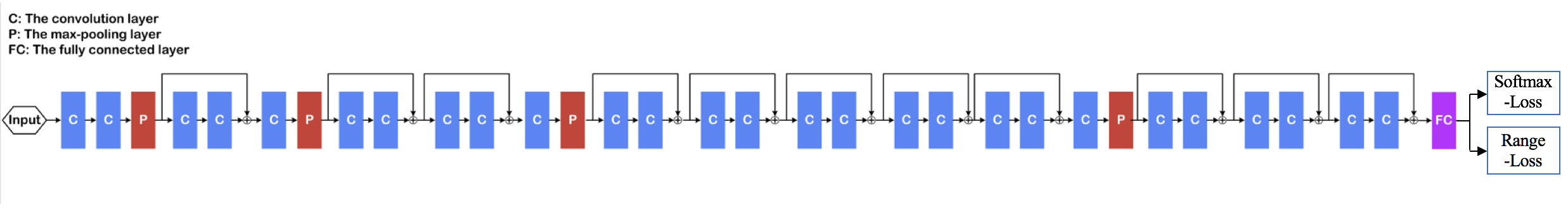}
\caption{Residual Network's structure adopted in our experiment. All the convolutional filters' size are 3$\times$3 with stride 1. Activation units ReLu layers are added after each convolutional layers. The number of the feature maps are 32 from the front layers to 512 in the last layers. We set the max-pooling's kernel size as 2$\times$2 with stride 2. Features in the last convolutional layer and the penultimate convolutional layer are extracted and concatenated as the input of the last fully connected layers. The whole CNN is trained under the joint supervisory signals of soft-max and our range loss.}
\label{fig:architecture}
\end{figure*}

\subsection{Implementation Details of VGG with Range Loss}
\textit{Training Data and Preprocessing:} To get a high-quality training data, we compute a mean feature vector for all identities according to their own pictures in data set. For a specific identity, images whose feature vector is far from the identity's feature vector will be removed. After carefully filtering and cleaning the MS-Celeb-1M \cite{guo2016ms} and CASIA- WebFace\cite{yi2014learning} data set, we obtain a dataset which contains 5M images with 100k unique identities. We use the new proposed multi-task cascaded CNN in \cite{zhang2016joint} to conduct the face detection and alignment. Training images are cropped to the size of 224$\times$224 and 112$\times$94 RGB 3-channel images for VGG and our CNN model's input, respectively. In this process, to estimate a reasonable mini-batch size is of crucial importance. By our experiences, it's better to construct such a mini-batch that contains multiple classes and same number of samples within each class. For examples, we set mini-batch size at 32 in our experiment, and 4 different identities in one batch with 8 images for each identity. For those small scale nets, it's normal to set 256 as the batch size, with 16 identities in one batch and 16 images per identities. Generally speaking, more identities being included in one mini-batch will contribute to both the softmax loss's supervising and the range loss's inter-class part.

\textit{VGG's settings:} The VGG net is a heavy convolutional neural networks model, especially when facing a training set with large amounts of identities. For 100k identities, according to our experiences, the mini-batch size can never exceed 32 because of the limitation of the GPU memory. The net is initialized by Gaussian distribution. The loss weight of the inter-class part of range loss is $10^{-4}$ while the intra-class part of range loss is $10^{-5}$. The parameter $margin$ is set $2\times10^4$. Initial learning rate is set at $0.1$ and reduce by half every $20,000$ iterations. We extract each of the testing sample's feature in the last fully connected layer.

\subsection{Performances on LFW and YTF Data sets}
LFW is a database of face photographs designed for unconstrained face recognition. The data set contains more than 13,000 images of faces collected from the web. Each face has been labeled with the name of the person pictured. 1680 of the people have two or more distinct photo's in this data set \cite{LFWTech}.

 YouTube faces database is a database of face videos designed for studying the problem of unconstrained face recognition in videos. The data set contains 3,425 videos of 1,595 different people. All the videos were downloaded from YouTube. An average of 2.15 videos are available for each subject \cite{wolf2011face}.
We implement the CNN model using the Caffe\cite{jia2014caffe} library with our customized range loss layers. For comparison, we train three models under the supervision of softmax loss (model A), joint contrastive loss and softmax loss (model B), and softmax combined with range loss (model C).
From the results shown in Table \ref{table:vgglfwytf}, we can see that Model C (jointly supervised by the range loss and softmax loss) beats the baseline model A (supervised by only softmax loss) by a large gap: from 97.87\% to 98.53\% in LFW. Contrary to our previous experimental result that models trained with complete long tailed data reach a lower accuracy, our model's (Model C) performances on complete long tail exceed the 50\% long tail group's result by 0.43\%. This shows that, firstly, comparing to soft-max loss and contrastive loss, range loss's capacity of learning discriminative feature from long tailed data performed best. Secondly, the integration of range loss to the model enables the latter 50\% tailed data to contribute to model's learning. This shows that, the original drawback that tailed data may bring, has been more than eliminated, but converted into notably contribution. This shows the advantage of our proposed range loss in dealing with long tailed data.

\begin{table}
\begin{center}
\begin{tabular}{|c|c|c|c|}
\hline
Methods &Tail Ratio&  LFW & YTF \\
\hline\hline
Model A& 50\%& \textbf{98.25\%}&\textbf{92.80\%}\\
Model A & 100\% &97.87\% &92.30\%\\
Model B   & 50\% &\textbf{98.47\%} &\textbf{93.20\%}\\
Model B  & 100\% &98.35\% &92.90\%\\
\hline
Model C & 50\% &98.45\% &93.20\%\\
Model C  & 100\% &\textbf{98.63\%} &\textbf{93.50\%}\\
\hline
\end{tabular}\\
\end{center}
\caption{Verification Accuracy of different loss combined with VGG on LFW and YTF data sets. Model A is using the softmax loss only. Model B is using the contrastive loss with softmax loss and Model C is using the range loss with softmax loss.}
\label{table:vgglfwytf}
\end{table}

\begin{table}
\begin{center}
\begin{tabular}{|c|c|c|c|}
\hline
Methods&Images &LFW & YTF \\
\hline\hline
DeepID-2+ \cite{sun2014deep}&   -  &99.47\%&93.20\%\\
FaceNet  \cite{schroff2015facenet}& 200M &99.63\%&95.10\%\\
Baidu   \cite{liu2015targeting} & 1.3M &99.13\%&-\\
Deep FR  \cite{parkhi2015deep}& 2.6M &98.95\%&97.30\%\\
DeepFace \cite{taigman2014deepface}& 4M   &97.35\%& 91.40\%\\
\hline
Model D &1.5M&98.27\% &93.10\% \\
Model E &1.5M&\textbf{99.52\%} &\textbf{93.70\%} \\
\hline
\end{tabular}\\
\end{center}
\caption{Compare Verification Accuracy of different CNN model on LFW and YTF datasets with our proposed CNN Networks. Model D is our adopted residual net with softmax loss only. Model E is the same net using range loss.}
\label{table:generaresult}
\end{table}

\subsection{Performance of Range Loss on other CNN structures}
To measure the performances and impact by the range loss and comprehensively and thoroughly, we further adopt residual CNN \cite{he2015deep} supervised by the joint signals of range loss and softmax. Deep residual net in recent years have been proved to show good generalization performance on recognition tasks. It
presents a residual learning framework that ease the training
of networks substantially deeper than those used
previously and up to 152 layers on the ImgageNet dataset.
That we choose this joint signals can be largely ascribed to the softmax's strong ability to give a discriminative boundaries among classes. Different to our previous practice, the model is trained under 1.5M filtered data from MS-Celeb-1M \cite{guo2016ms} and CASIA- WebFace\cite{yi2014learning}, which is of smaller scale size of the original long tail dataset with a more uniform distribution. The intention of this experiment lies that: apart from the ability to utilize amounts of imbalanced data, we want to verify our loss function's  generalization ability to train universal CNN model and to achieve the state-of-art performances. We evaluate the range loss based residual net's performances on LFW and YTF's face verification task.
The model's architecture is illustrated in Fig.\ref{fig:architecture}.
In Table \ref{table:generaresult}, we compare our method against many existing models, including DeepID-2+\cite{sun2014deep}, FaceNet\cite{schroff2015facenet}, Baidu\cite{liu2015targeting}, DeepFace\cite{taigman2014deepface} and our baseline model D (Our residual net structure supervised by softmax loss). From the results in Table \ref{table:generaresult}, we have the following observations. Firstly, our model E (supervised by softmax and range loss) beats the
baseline model D (supervised by softmax only) by a significant margin (from 98.27\% to 99.52\% in LFW, and 93.10\% to 93.70\% in YTF). This represents the joint supervision of range loss and softmax loss can notablely enhance the deep neural models' ability to extract discriminative features. Secondly, residual network integrated with range loss was non-inferior to the existing famous networks and even outperforms most of them. This shows our loss function's  generalization ability to train universal CNN model and to achieve the state-of-art performances. Lastly, our proposed networks are trained under a database far less than other's(shown in Table \ref{table:generaresult}), this indicates the advantages of our  network.

\section{Conclusions}
In this paper, we deeply explore the potential effects the long tail distribution may pose to the deep
model’s training. Contrary to our intuitiveness, long tailed data, if tailored properly, can contribute to the model's training. We proposed a new loss function, namely range loss. By combining the range loss with the softmax loss to jointly supervise the learning of CNNs, it is able to reduce the intra-class variations and enlarge the inter-class distance under imbalanced long tailed data effectively. Therefore, the optimization goal towards the poor classes should be focused on these thorny samples within one class.
 Its performance on several large-scale face benchmarks has convincingly demonstrated the effectiveness of the proposed approach.

{\small
\bibliographystyle{ieee}
\bibliography{egbib}

\begin{thebibliography}{10}\itemsep=-1pt

\bibitem{Bingham2011The}
A.~Bingham and D.~Spradlin.
\newblock The long tail of expertise.
\newblock 2011.

\bibitem{chen2013blessing}
D.~Chen, X.~Cao, F.~Wen, and J.~Sun.
\newblock Blessing of dimensionality: High-dimensional feature and its
  efficient compression for face verification.
\newblock In {\em Proceedings of the IEEE Conference on Computer Vision and
  Pattern Recognition}, pages 3025--3032, 2013.

\bibitem{deng2009imagenet}
J.~Deng, W.~Dong, R.~Socher, L.-J. Li, K.~Li, and L.~Fei-Fei.
\newblock Imagenet: A large-scale hierarchical image database.
\newblock In {\em Computer Vision and Pattern Recognition, 2009. CVPR 2009.
  IEEE Conference on}, pages 248--255. IEEE, 2009.

\bibitem{Everingham10}
M.~Everingham, L.~Van~Gool, C.~K.~I. Williams, J.~Winn, and A.~Zisserman.
\newblock The pascal visual object classes (voc) challenge.
\newblock {\em International Journal of Computer Vision}, 88(2):303--338, June
  2010.

\bibitem{griffin2007caltech}
G.~Griffin, A.~Holub, and P.~Perona.
\newblock Caltech-256 object category dataset.
\newblock 2007.

\bibitem{guo2016ms}
Y.~Guo, L.~Zhang, Y.~Hu, X.~He, and J.~Gao.
\newblock Ms-celeb-1m: A dataset and benchmark for large-scale face
  recognition.
\newblock In {\em European Conference on Computer Vision}, pages 87--102.
  Springer, 2016.

\bibitem{gupta2014learning}
S.~Gupta, R.~Girshick, P.~Arbel{\'a}ez, and J.~Malik.
\newblock Learning rich features from rgb-d images for object detection and
  segmentation.
\newblock In {\em European Conference on Computer Vision}, pages 345--360.
  Springer, 2014.

\bibitem{he2014spatial}
K.~He, X.~Zhang, S.~Ren, and J.~Sun.
\newblock Spatial pyramid pooling in deep convolutional networks for visual
  recognition.
\newblock In {\em European Conference on Computer Vision}, pages 346--361.
  Springer, 2014.

\bibitem{he2015deep}
K.~He, X.~Zhang, S.~Ren, and J.~Sun.
\newblock Deep residual learning for image recognition.
\newblock {\em arXiv preprint arXiv:1512.03385}, 2015.

\bibitem{he2015delving}
K.~He, X.~Zhang, S.~Ren, and J.~Sun.
\newblock Delving deep into rectifiers: Surpassing human-level performance on
  imagenet classification.
\newblock In {\em Proceedings of the IEEE International Conference on Computer
  Vision}, pages 1026--1034, 2015.

\bibitem{henriques2013beyond}
J.~F. Henriques, J.~Carreira, R.~Caseiro, and J.~Batista.
\newblock Beyond hard negative mining: Efficient detector learning via
  block-circulant decomposition.
\newblock In {\em proceedings of the IEEE International Conference on Computer
  Vision}, pages 2760--2767, 2013.

\bibitem{LFWTech}
G.~B. Huang, M.~Ramesh, T.~Berg, and E.~Learned-Miller.
\newblock Labeled faces in the wild: A database for studying face recognition
  in unconstrained environments.
\newblock Technical Report 07-49, University of Massachusetts, Amherst, October
  2007.

\bibitem{jia2014caffe}
Y.~Jia, E.~Shelhamer, J.~Donahue, S.~Karayev, J.~Long, R.~Girshick,
  S.~Guadarrama, and T.~Darrell.
\newblock Caffe: Convolutional architecture for fast feature embedding.
\newblock {\em arXiv preprint arXiv:1408.5093}, 2014.

\bibitem{karpathy2014large}
A.~Karpathy, G.~Toderici, S.~Shetty, T.~Leung, R.~Sukthankar, and L.~Fei-Fei.
\newblock Large-scale video classification with convolutional neural networks.
\newblock In {\em Proceedings of the IEEE conference on Computer Vision and
  Pattern Recognition}, pages 1725--1732, 2014.

\bibitem{krizhevsky2012imagenet}
A.~Krizhevsky, I.~Sutskever, and G.~E. Hinton.
\newblock Imagenet classification with deep convolutional neural networks.
\newblock In {\em Advances in neural information processing systems}, pages
  1097--1105, 2012.

\bibitem{lin2013network}
M.~Lin, Q.~Chen, and S.~Yan.
\newblock Network in network.
\newblock {\em arXiv preprint arXiv:1312.4400}, 2013.

\bibitem{liu2015targeting}
J.~Liu, Y.~Deng, and C.~Huang.
\newblock Targeting ultimate accuracy: Face recognition via deep embedding.
\newblock {\em arXiv preprint arXiv:1506.07310}, 2015.

\bibitem{norouzi2013zero}
M.~Norouzi, T.~Mikolov, S.~Bengio, Y.~Singer, J.~Shlens, A.~Frome, G.~S.
  Corrado, and J.~Dean.
\newblock Zero-shot learning by convex combination of semantic embeddings.
\newblock {\em arXiv preprint arXiv:1312.5650}, 2013.

\bibitem{ouyang2016factors}
W.~Ouyang, X.~Wang, C.~Zhang, and X.~Yang.
\newblock Factors in finetuning deep model for object detection with long-tail
  distribution.
\newblock In {\em Proceedings of the IEEE Conference on Computer Vision and
  Pattern Recognition}, pages 864--873, 2016.

\bibitem{parkhi2015deep}
O.~M. Parkhi, A.~Vedaldi, and A.~Zisserman.
\newblock Deep face recognition.
\newblock In {\em British Machine Vision Conference}, volume~1, page~6, 2015.

\bibitem{ILSVRC15}
O.~Russakovsky, J.~Deng, H.~Su, J.~Krause, S.~Satheesh, S.~Ma, Z.~Huang,
  A.~Karpathy, A.~Khosla, M.~Bernstein, A.~C. Berg, and L.~Fei-Fei.
\newblock {ImageNet Large Scale Visual Recognition Challenge}.
\newblock {\em International Journal of Computer Vision (IJCV)},
  115(3):211--252, 2015.

\bibitem{againstlongtail}
S.Bengio.
\newblock The battle against the long tail.
\newblock Talk on Workshop on Big Data and Statistical Machine Learning., 2015.

\bibitem{schroff2015facenet}
F.~Schroff, D.~Kalenichenko, and J.~Philbin.
\newblock Facenet: A unified embedding for face recognition and clustering.
\newblock In {\em Proceedings of the IEEE Conference on Computer Vision and
  Pattern Recognition}, pages 815--823, 2015.

\bibitem{sermanet2013overfeat}
P.~Sermanet, D.~Eigen, X.~Zhang, M.~Mathieu, R.~Fergus, and Y.~LeCun.
\newblock Overfeat: Integrated recognition, localization and detection using
  convolutional networks.
\newblock {\em arXiv preprint arXiv:1312.6229}, 2013.

\bibitem{simonyan2014very}
K.~Simonyan and A.~Zisserman.
\newblock Very deep convolutional networks for large-scale image recognition.
\newblock {\em arXiv preprint arXiv:1409.1556}, 2014.

\bibitem{sun2014deep}
Y.~Sun, Y.~Chen, X.~Wang, and X.~Tang.
\newblock Deep learning face representation by joint
  identification-verification.
\newblock In {\em Advances in Neural Information Processing Systems}, pages
  1988--1996, 2014.

\bibitem{szegedy2015going}
C.~Szegedy, W.~Liu, Y.~Jia, P.~Sermanet, S.~Reed, D.~Anguelov, D.~Erhan,
  V.~Vanhoucke, and A.~Rabinovich.
\newblock Going deeper with convolutions.
\newblock In {\em Proceedings of the IEEE Conference on Computer Vision and
  Pattern Recognition}, pages 1--9, 2015.

\bibitem{taigman2014deepface}
Y.~Taigman, M.~Yang, M.~Ranzato, and L.~Wolf.
\newblock Deepface: Closing the gap to human-level performance in face
  verification.
\newblock In {\em Proceedings of the IEEE Conference on Computer Vision and
  Pattern Recognition}, pages 1701--1708, 2014.

\bibitem{wen2016latent}
Y.~Wen, Z.~Li, and Y.~Qiao.
\newblock Latent factor guided convolutional neural networks for age-invariant
  face recognition.
\newblock In {\em Proceedings of the IEEE Conference on Computer Vision and
  Pattern Recognition}, pages 4893--4901, 2016.

\bibitem{wen2016discriminative}
Y.~Wen, K.~Zhang, Z.~Li, and Y.~Qiao.
\newblock A discriminative feature learning approach for deep face recognition.
\newblock In {\em European Conference on Computer Vision}, pages 499--515.
  Springer, 2016.

\bibitem{wolf2011face}
L.~Wolf, T.~Hassner, and I.~Maoz.
\newblock Face recognition in unconstrained videos with matched background
  similarity.
\newblock In {\em Computer Vision and Pattern Recognition (CVPR), 2011 IEEE
  Conference on}, pages 529--534. IEEE, 2011.

\bibitem{yang2014context}
J.~Yang, B.~Price, S.~Cohen, and M.-H. Yang.
\newblock Context driven scene parsing with attention to rare classes.
\newblock In {\em Proceedings of the IEEE Conference on Computer Vision and
  Pattern Recognition}, pages 3294--3301, 2014.

\bibitem{yi2014learning}
D.~Yi, Z.~Lei, S.~Liao, and S.~Z. Li.
\newblock Learning face representation from scratch.
\newblock {\em arXiv preprint arXiv:1411.7923}, 2014.

\bibitem{zhang2016joint}
K.~Zhang, Z.~Zhang, Z.~Li, and Y.~Qiao.
\newblock Joint face detection and alignment using multi-task cascaded
  convolutional networks.
\newblock {\em arXiv preprint arXiv:1604.02878}, 2016.

\bibitem{zhou2014object}
B.~Zhou, A.~Khosla, A.~Lapedriza, A.~Oliva, and A.~Torralba.
\newblock Object detectors emerge in deep scene cnns.
\newblock {\em arXiv preprint arXiv:1412.6856}, 2014.

\bibitem{zhou2014learning}
B.~Zhou, A.~Lapedriza, J.~Xiao, A.~Torralba, and A.~Oliva.
\newblock Learning deep features for scene recognition using places database.
\newblock In {\em Advances in neural information processing systems}, pages
  487--495, 2014.

\bibitem{zhou2015naive}
E.~Zhou, Z.~Cao, and Q.~Yin.
\newblock Naive-deep face recognition: Touching the limit of lfw benchmark or
  not?
\newblock {\em arXiv preprint arXiv:1501.04690}, 2015.

\end{thebibliography}
}

\end{document}